\documentclass[10pt, a4paper]{article}
\usepackage[]{lrec-coling2024}

\usepackage{graphicx}
\usepackage{tabularx}
\usepackage{soul}
\usepackage{float}

\usepackage{xcolor}
\usepackage{hyperref}
 \definecolor{darkblue}{rgb}{0, 0, 0.5}
  \hypersetup{colorlinks=true, citecolor=darkblue, linkcolor=darkblue, urlcolor=darkblue}

\usepackage{xstring}

\usepackage{color}

\usepackage{todonotes}
\usepackage{booktabs}

\definecolor{cadmiumgreen}{rgb}{0.0, 0.42, 0.24}
\definecolor{cadmiumred}{rgb}{0.89, 0.0, 0.13}
\definecolor{cadmiumorange}{rgb}{0.93, 0.53, 0.18}

\newcommand{\tabspace}{\hspace{0.15cm}}
\newcommand{\tspace}{\hspace{0.2cm}}
\newcommand{\tred}[1]{\textcolor{cadmiumred}{\textbf{#1}}}
\newcommand{\tgreen}[1]{\textcolor{cadmiumgreen}{\textbf{#1}}}

\newcommand{\para}[1]{\paragraph{#1}}

\title{Do Language Models Care About Text Quality?\\Evaluating Web-Crawled Corpora Across 11 Languages}

\name{Rik van Noord\textsuperscript{$\diamondsuit$}, Taja Kuzman\textsuperscript{$\clubsuit$}, Peter Rupnik\textsuperscript{$\clubsuit$}, Nikola Ljubešić\textsuperscript{$\clubsuit$}\\\large \textbf{Miquel Esplà-Gomis\textsuperscript{$\spadesuit$}, Gema Ramírez-Sánchez\textsuperscript{$\heartsuit$} and Antonio Toral\textsuperscript{$\diamondsuit$}}\vspace{0.05cm}} 

\address{\textsuperscript{$\diamondsuit$}University of Groningen, \textsuperscript{$\clubsuit$}Jožef Stefan Institute, \textsuperscript{$\spadesuit$}Universitat d'Alacant, \textsuperscript{$\heartsuit$}Prompsit\\
         rikvannoord@gmail.com
         \vspace{-0.05cm}\\}

\abstract{
\vspace{0.1cm}
Large, curated, web-crawled corpora play a vital role in training language models (LMs). They form the lion's share of the training data in virtually all recent LMs, such as the well-known GPT, LLaMA and XLM-RoBERTa models. However, despite this importance, relatively little attention has been given to the quality of these corpora. In this paper, we compare four of the currently most relevant large, web-crawled corpora (CC100, MaCoCu, mC4 and OSCAR) across eleven lower-resourced European languages. Our approach is two-fold: first, we perform an intrinsic evaluation by performing a human evaluation of the quality of samples taken from different corpora; then, we assess the practical impact of the qualitative differences by training specific LMs on each of the corpora and evaluating their performance on downstream tasks. We find that there are clear differences in \emph{quality} of the corpora, with MaCoCu and OSCAR obtaining the best results. However, during the extrinsic evaluation, we actually find that the CC100 corpus achieves the highest scores. We conclude that, in our experiments, the quality of the web-crawled corpora does not seem to play a significant role when training LMs.
 \\  
 \newline  \vspace{0.1cm} }

\begin{document}

\maketitleabstract

\vspace{0.1cm}
\section{Introduction}
The field of natural language processing has witnessed a paradigm shift with the emergence of large language models (LLMs) that exhibit impressive capabilities in various language understanding tasks \citep{OPT, GPT4, LLAMA}. Monolingual corpora have played a pivotal role in this data-driven revolution, serving as the foundational resource for training these LLMs. A growing number of monolingual corpora have been published in the last years, many of them specifically conceived to train LLMs~\citep{xlmr, mc4}, by implementing different methodologies to collect and curate data. However, despite their importance, the content of these corpora has been given modest attention. Since these corpora are compiled using automatic tools, with limited and varying quality control, it is unclear (i) how these corpora are qualitatively different and (ii) if and how the differences actually affect the models in terms of downstream performance.

In this paper, we aim to shed light on these issues. In particular, we evaluate the well-known OSCAR~\citep{oscar}, CC100~\citep{xlmr}, mC4~\citep{mc4} and MaCoCu~\citep{macocu} corpora across eleven non-English European languages. First, we hire professional linguists to manually evaluate the \emph{quality} of the corpora, irrespective of the size. Then, we train a language model on each language-corpus combination for a subset of five languages, to evaluate whether these differences in quality transfer to differences in downstream performance. All code, models and annotations are made publicly available.\footnote{\url{https://github.com/RikVN/Corpus_Eval/}}

In terms of quality, our findings indicate that the MaCoCu and OSCAR corpora are superior options.  They contain a greater number of documents that consist of (publishable) running text, while exhibiting a significantly lower number of documents that are either in an incorrect language or lack running text. The mC4 corpus seems to be the one of the lowest quality in our evaluation, with especially concerning results for Maltese, as over 75\% of the corpus was in another language.

However, despite the evident differences in quality, the findings do not seem to directly transfer to actually training the LMs on the corpora in question. For a subset of five languages, we continue training XLM-RoBERTa (XLM-R, \citealp{xlmr}) on each of our corpora and evaluate performance across a number of structured prediction and natural language understanding tasks. We find that CC100 actually obtains the best performance, while OSCAR is the worst performing corpus. We intentionally refrained from controlling for data set size, as the size of the data sets is associated with the extent of data cleaning conducted, which is undeniably linked to the quality of the preserved texts. However, even if we do control for size, we do not find any indication that the quality of the data significantly influences performance, counter-intuitive as it may be.

\section{Related work}

Given the important role web-crawled corpora play in training LLMs, and given that they are known to be noisy \citep{junczys-dowmunt-2019-microsoft,luccioni-viviano-2021-whats}, it is curious that there are only a few papers that actually aimed to evaluate the quality of these corpora. \citet{caswell-etal-2020-language} analysed the performance of their automatic language identification system and found serious issues for lower-resource languages. \citet{dodge-etal-2021-documenting} documented the C4 data set used for training mC4 \citep{mc4} and found that a significant number of texts came from unexpected sources, such as US military websites. Moreover, they found a substantial amount of machine-generated text and texts from common NLP evaluation benchmarks.
Closer to our work is the study by \citet{kreutzer-etal-2022-quality}, in which they run a human evaluation on a large number of both monolingual and parallel corpora. 
The authors focus mostly on the languages for which less data is available in each corpus, but they also include other languages with more data for a more representative evaluation. They find serious quality issues for the evaluated corpora, especially for low resource languages, but do not run an automatic evaluation.

\vspace{-0.1cm}
\para{Basque} The
closest
to our work is the study of \citet{artetxe-etal-2022-corpus}. They perform both 
human and automatic evaluation on monolingual corpora for Basque, where the automatic evaluation is extrinsic, 
by training LMs on these corpora and then evaluating them on several downstream tasks.
They conclude that there is no clear correlation between either the size or the quality of data and the performance of the LMs trained on them. Our work follows~\citet{artetxe-etal-2022-corpus}, also evaluating the \emph{quality} of monolingual corpora and the performance of LMs trained on them, but we do so for several languages and include a larger number of web crawled corpora. 

\vspace{-0.1cm}
\para{Cleaning} It should be noted that the corpora under evaluation have undergone a preliminary filtering and cleaning process prior to their release (see Section~\ref{sec:corpora}). However, these (or similar) corpora often go through an additional cleaning process before being used to train an LM \citep{rae2021scaling,LLAMA,GPT4}. However, these cleaning practices are all (slightly) different and are usually not explained in enough detail to ensure reproducibility. Since we have no access to each individual cleaning process, in this paper, we explicitly \textbf{do not} attempt to find the best cleaning methods. We simply evaluate existing monolingual corpora by using them \emph{as is} and only performing the simple cleaning steps as instructed by their creators.

\para{Monolingual LMs} In our automatic evaluation, we train encoder-only monolingual language models. Even though many current studies focus on large decoder-only models, we believe there is still a need for smaller, monolingual LMs. This is evidenced by the use of such models to \emph{enrich} corpora at scale in a computationally-efficient manner \citep{kuzman2023automatic}, but also simply by the popularity of such models \citep{bertje,sanh2019distilbert,camembert,flaubert,souza2020bertimbau,berturk,ljubesic-lauc-2021-bertic,icebert,seker-etal-2022-alephbert} on the HuggingFace hub.\footnote{For example, CamemBERT \citep{camembert} had over 2.5 million downloads last month (as of March 2024).}

\para{Continued training} In this paper, we continue training an existing LM, instead of training from scratch. The main advantage of this approach is that it is a lot more efficient, while results remain competitive or even improve \citep{gururangan-etal-2020-dont, wang2020extending,chau2020parsing,muller2021being,icebert}. A similar option is \emph{adapting} LMs to a target language \citep{pfeiffer2020mad,pfeiffer2021adapterfusion} by learning language-specific representations. However, \citet{ebrahimi2021adapt} found that continued pretraining provided the best results on low-resource languages, while also being the
simplest
method to apply, which is why we use this method in our paper.

\section{Corpora\label{sec:corpora}}

Four corpora were included in the evaluation described in this work. In this section, we provide a general overview of each of them.

\para{CC100~\citep{xlmr}.} Corpus created  
to train the popular XLM-R language model~\citep{xlmr}. The corpus covers 100 languages 
and was built through cleaning twelve Common Crawl dumps,\footnote{\url{https://data.commoncrawl.org/}} using one of them to extract only text in English, and the remaining eleven dumps to extract data in the rest of languages. The per-language size of the corpus ranges from 55.6 billion
tokens in English 
to 10 million tokens in Sundanese. 
This corpus was built by following the approach of~\citet{wenzek-etal-2020-ccnet}, which consists of three main steps: (a) deduplication of paragraphs on Common Crawl dumps; (b) language identification with fastText~\citep{grave-etal-2018-learning}
and (c) providing, for each paragraph in the corpus, the perplexity as provided by a language model as a proxy of the quality of the text. 

\para{mC4~\citep{mc4}.} Corpus created to build the mT5 language model~\citep{mc4}. The corpus covers 101 languages
and was built by processing all the Common Crawl dumps available at that time. The tool cld3\footnote{\url{https://github.com/google/cld3}} was used for language identification, deduplication was performed at paragraph level and pages with too few or too short paragraphs or with bad words were filtered out. 
The final size of the mC4 corpus is approximately 6.3 trillion tokens in total. 

\vspace{-0.15cm}
\para{OSCAR~\citep{oscar}.} This corpus is also built through a cleaning and curation process on Common Crawl data. It is developed incrementally, with regular releases of new versions including data from the last Common Crawl dumps.
According to the documentation of the last version of the corpus at the time of writing 
(v23/01),\footnote{\url{https://oscar-project.github.io/documentation/versions/oscar-2301/}} OSCAR uses fastText~\citep{grave-etal-2018-learning} for language identification and the TLSH~\citep{TLSH} fuzzy hashing method to identify near duplicates. This version of OSCAR also provides the perplexity provided by a KenLM~\citep{heafield-2011-kenlm} language model trained on harmful content identified in previous versions of OSCAR.
OSCAR is the corpus that covers the most languages among the collections evaluated in this paper, with a total of 152 languages in its latest version. 
However, it is worth noting that for about 50 of them, the corpus includes less than 1 million tokens.\footnote{Some languages have extremely little data, such as Kalmyk (27 tokens) or Quechua (13 tokens).}

\vspace{-0.15cm}
\paragraph{MaCoCu~\citep{macocu}.} In contrast to the rest of corpora compared in this paper, MaCoCu corpora are not obtained by processing Common Crawl data. 
Instead, a strategy consisting of crawling relevant internet top-level domains directly
for the targeted languages is followed (e.g., .al for Albanian). Several studies claim that Common Crawl over-represents English while under-represents other languages~\citep{bender2021,ranathunga-de-silva-2022-languages}; the strategy adopted is aimed at avoiding this effect and, at the same time, granting access to more up-to-date data than that stored in Common Crawls.
The MaCoCu corpus covers 11 low-resourced European languages, and consists of a total of about 17.3 billion tokens, with Turkish being the largest (4.3 billion tokens) and Montenegrin being the smallest (200 million tokens).
During cleaning, deduplication and a set of heuristics is applied to fix or remove evidently problematic text fragments, such as badly encoded or too short ones.

\section{Manual Evaluation}

In this section, we describe the methodology and results of the manual evaluation of the CC100, MaCoCu, mC4 and OSCAR corpora. We evaluate all languages that are present in MaCoCu, the corpus with the smallest amount of languages present. It should be noted that certain languages found in the MaCoCu corpus are exclusive to this particular corpus and are not represented in any other corpora, namely Bosnian and Montenegrin. While we include these languages in the human evaluation,  we mostly focus on the  following nine languages that are present in multiple corpora: Albanian, Bulgarian, Croatian, Icelandic, Macedonian, Maltese, Serbian, Slovenian and Turkish.

\subsection{Annotation Scheme}

We perform annotation at paragraph level as this is the common format that each corpus has available. For OSCAR, we follow the best practice standard of only selecting the paragraphs that are recognized as being in the correct language.
The other corpora are used as 
they are released. For each corpus and language combination, we randomly select 200 paragraphs for annotation. Annotators are asked to rank each paragraph using the following scale:

\begin{enumerate}
    \itemsep0em 
    \item \textbf{Wrong language or not language (WL).} The text is not in the correct language, or is not in a natural language (e.g., links, html tags).
    \item \textbf{Not running text (NR).} The text makes no sense, it is just a concatenation of words or a bunch of words together. Note that short sentences can still be running text.
    \item \textbf{Partially running text (PR).} More than 50\% is running text, but some parts are not. For example, the text is cut-off or has additional elements in brackets. A substantial part of the text should be cut-off for this to apply.
    \item \textbf{Running text, but slightly non-standard (RT).} More than 90\% is running text, but the text contains small mistakes, such as grammatical errors, typos, and missing punctuation. This category includes titles, headers and bullet points.
    \item \textbf{Publishable text (PT).} 100\% running text which is of publishable quality and contains no (formatting) mistakes. You could read this in a blog post, news article, recipe, magazine, etc. Note that the content itself does not have to be formal for this to apply.
\end{enumerate}

This schema is inspired by those in two recent works \cite{kreutzer-etal-2022-quality, artetxe-etal-2022-corpus}, which were taken as a starting point, combined, and refined by means of a pilot annotation conducted on Slovenian and English samples. Each annotator is also provided with a number of example paragraphs in English\footnote{All our annotators were also fluent in English. The advantage of using this language as instruction is that the instructions are the same across all languages.}, for each category. These are compiled in Table~\ref{tab:examples_mono} of the Appendix. 
The annotators are instructed to select only one option, and to pick the lower number on the scale in case of uncertainty.

\begin{table}[!t]
    \centering
    \begin{tabular}{l|cc}
         \toprule
                     & \multicolumn{1}{c}{\textbf{\% agreement}} & \textbf{$\kappa$ coef.} \\
         \midrule
\textbf{Albanian}    & \multicolumn{1}{c|}{68.0}        & 0.51              \\
\textbf{Bosnian}     & \multicolumn{1}{c|}{86.5}        & 0.59              \\
\textbf{Bulgarian}   & \multicolumn{1}{c|}{49.5}        & 0.32              \\
\textbf{Croatian}    & \multicolumn{1}{c|}{72.1}        & 0.62              \\
\textbf{Icelandic}   & \multicolumn{1}{c|}{81.0}        & 0.39              \\
\textbf{Macedonian}  & \multicolumn{1}{c|}{48.5}        & 0.27              \\
\textbf{Maltese}     & \multicolumn{1}{c|}{66.1}        & 0.55              \\
\textbf{Montenegrin} & \multicolumn{1}{c|}{47.5}        & 0.23              \\
\textbf{Serbian}     & \multicolumn{1}{c|}{78.0}        & 0.65              \\
\textbf{Slovenian}   & \multicolumn{1}{c|}{70.5}        & 0.36              \\
\textbf{Turkish}     & \multicolumn{1}{c|}{52.5}        & 0.32             \\
         \bottomrule
    \end{tabular}
    \vspace{0.5cm}
    \caption{\label{tab:inter}Inter-annotator agreement between the two annotators for each language for the evaluation of monolingual data. The second column shows the percentage (\textbf{\%}) of annotations for which both annotators were in exact agreement; the third column shows Cohen's kappa coefficient (\textbf{$\kappa$}) between both annotators.} 
\end{table}

\para{Details} We hire two professional linguists per language to annotate subsets of all corpora in each of the 11 languages. The evaluation samples comprise 200 instances from each evaluated corpus. The size of the samples depends on the number of corpora in which the evaluated language is present. Specifically, the size ranges from 200 instances when only one corpus is available to 800 instances when all four corpora are included in the comparison. From the sample, we select 200 instances that are provided to both annotators 
to be able to calculate inter-annotator agreement. We balance the instances for each corpus per annotator, meaning that each annotator sees 100 instances of each corpus. The instances are shown to them in random order and blind fashion, i.e., they do not know which instance belongs to which corpus. For the 200 instances that have double annotations, we select one of them randomly to be included in the analysis (balanced per annotator). The annotators received a fair wage
for their efforts that differed per language, but always exceeded minimum wage.

\begin{table}[H]
    \centering
        \setlength{\tabcolsep}{4pt}
     \resizebox{\columnwidth}{!}{
    \begin{tabular}{lrrrrrr}
         \toprule
         \textbf{Albanian} & \textbf{WL} & \textbf{NR} & \textbf{PR} & \textbf{RT} & \textbf{PT}  & \textbf{RT+PT} \\
            \tspace MaCoCu & 4   & 4   & 48  & 73  & 71  & 144   \\
            \tspace CC100    & 1   & 3   & 44  & 62  & 90  & 152  \\
            \tspace mC4      & 18  & 12  & 58  & 47  & 65  & \tred{112}   \\
            \tspace OSCAR    & 1   & 2   & 32  & 70  & 95  & \tgreen{165}  \\
            \midrule
            \textbf{Bosnian} & \textbf{WL} & \textbf{NR} & \textbf{PR} & \textbf{RT} & \textbf{PT}  & \textbf{RT+PT} \\
            \tspace MaCoCu & 2   & 0   & 3   & 38  & 157 & \tgreen{195}    \\
            \midrule
            \textbf{Bulgarian} & \textbf{WL} & \textbf{NR} & \textbf{PR} & \textbf{RT} & \textbf{PT}  & \textbf{RT+PT} \\
            \tspace MaCoCu & 5   & 8   & 18  & 78  & 91  & \tgreen{169}   \\
            \tspace CC100    & 2   & 23  & 18  & 66  & 91  & 157   \\
             \tspace mC4      & 17  & 49  & 25  & 47  & 62  & \tred{109}   \\
             \tspace OSCAR    & 2   & 26  & 26  & 58  & 88  & 146   \\
            \midrule
           \textbf{Croatian} & \textbf{WL} & \textbf{NR} & \textbf{PR} & \textbf{RT} & \textbf{PT}  & \textbf{RT+PT} \\
            \tspace MaCoCu & 10  & 23  & 23  & 61  & 83  & \tgreen{144}   \\
            \tspace CC100    & 18  & 15  & 25  & 71  & 71  & 142   \\
            \tspace OSCAR    & 1   & 37  & 32  & 65  & 64  & \tred{129}   \\
            \midrule
            \textbf{Icelandic} & \textbf{WL} & \textbf{NR} & \textbf{PR} & \textbf{RT} & \textbf{PT}  & \textbf{RT+PT} \\
            \tspace MaCoCu & 2   & 4   & 6   & 15  & 173 & 188    \\ 
            \tspace CC100    & 2   & 6   & 9   & 19  & 164 & 183    \\ 
            \tspace mC4      & 24  & 15  & 16  & 15  & 130 & \tred{145}    \\
           \tspace  OSCAR    & 2   & 1   & 4   & 4   & 189 & \tgreen{193}   \\
            \midrule
            \textbf{Macedonian} & \textbf{WL} & \textbf{NR} & \textbf{PR} & \textbf{RT} & \textbf{PT}  & \textbf{RT+PT} \\
            \tspace MaCoCu & 5 &     5 &    26 &    76 &   88 &     \tgreen{164}    \\
            \tspace CC100    & 1 &    11 &    41 &    76 &   71 &     147 \\
            \tspace mC4      & 10 &    20 &    30 &    59 &   81 &     \tred{140}   \\
            \tspace OSCAR    & 2 &     7 &    31 &    64 &   96 &     160   \\
              \midrule
              \textbf{Maltese} & \textbf{WL} & \textbf{NR} & \textbf{PR} & \textbf{RT} & \textbf{PT}  & \textbf{RT+PT} \\
              \tspace MaCoCu & 9   & 101 & 38  & 16  & 36  & 52    \\
              \tspace mC4      & 164 & 17  & 4   & 1   & 14  & \tred{15}    \\
              \tspace OSCAR    & 7   & 32  & 30  & 17  & 98  & \tgreen{115}   \\
            \midrule
            \textbf{Montenegrin} & \textbf{WL} & \textbf{NR} & \textbf{PR} & \textbf{RT} & \textbf{PT}  & \textbf{RT+PT} \\
            \tspace MaCoCu & 25  & 4   & 18  & 49  & 104 & \tgreen{153} \\
            \midrule
            \textbf{Serbian} & \textbf{WL} & \textbf{NR} & \textbf{PR} & \textbf{RT} & \textbf{PT}  & \textbf{RT+PT} \\
            \tspace MaCoCu & 1   & 1   & 14  & 82  & 102 & \tgreen{184}   \\
            \tspace CC100    & 0   & 5   & 24  & 60  & 111 & 171   \\
            \tspace mC4      & 5   & 14  & 47  & 65  & 69  & \tred{134}   \\
            \tspace OSCAR    & 0   & 1   & 24  & 69  & 106 & 175   \\
            \midrule
             \textbf{Slovenian}   & \textbf{WL} & \textbf{NR} & \textbf{PR} & \textbf{RT} & \textbf{PT}  & \textbf{RT+PT} \\
             \tspace MaCoCu  &     3 &     4 &    14 &    33 &  146 &  179 \\
             \tspace CC100     &     1 &    13 &    29 &    15 &  142 &  157 \\
             \tspace mC4       &    11 &    22 &    23 &    30 &  114 &     \tred{144} \\
             \tspace OSCAR     &     0 &     2 &    12 &    13 &  173 &     \tgreen{186} \\
             \midrule
             \textbf{Turkish} & \textbf{WL} & \textbf{NR} & \textbf{PR} & \textbf{RT} & \textbf{PT}  & \textbf{RT+PT} \\
             \tspace MaCoCu & 5   & 27  & 19  & 90  & 59  & \tgreen{149}   \\
             \tspace CC100    & 0   & 33  & 30  & 95  & 42  & 137   \\
             \tspace mC4      & 8   & 62  & 30  & 67  & 33  & \tred{100}   \\
             \tspace OSCAR    & 3   & 38  & 26  & 84  & 49  & 133  \\
                     \bottomrule
    \end{tabular}
    }
    \vspace{0.1cm}
    \caption{\label{tab:human_mono}Summary of the human evaluation of the web-crawled corpora. Paragraphs were annotated as Wrong Language (WL), Not-running text (NR), Partially Running Text (PR), Running Text (RT) or Publishable Text (PT).
    }
\end{table}

\vspace{-0.1cm}
\subsection{Annotation Results}
\vspace{-0.1cm}

The inter-annotator agreement in terms of exact annotation overlap and Cohen's kappa coefficient ($\kappa$) scores are shown in Table~\ref{tab:inter}. Generally, the annotators agree a fair amount of the time. It is not surprising that there is some disagreement: for example, the difference between ``Running Text'' and ``Publishable Text'' is partially subjective. These are indeed the two annotation categories that annotators disagree most often about, as can be seen in Table~\ref{tab:cm}. The instances where annotators disagree about a paragraph being in the wrong language generally come exclusively from the Serbo-Croatian languages (Bosnian, Croatian, Montenegrin and Serbian).

\begin{table}[!tb]
    \centering
    \begin{tabular}{l|rrrrr}
         \toprule
       & \textbf{WL} & \textbf{NR} & \textbf{PR} & \textbf{RT} & \textbf{PT} \\
          \midrule
       \textbf{WL} & \textbf{101}  &  --- &  --- &  --- &  --- \\
        \textbf{NR} & 29  & \textbf{94}  &  --- &  --- &  ---\\
         \textbf{PR} & 10  & 56  & \textbf{108} &  --- &  ---\\
          \textbf{RT}  & 14  & 38  & 117  & \textbf{275}  &  ---\\
           \textbf{PT}  & 8  & 32 & 77  & 371  & \textbf{847}  \\
         \bottomrule
    \end{tabular}
    \caption{\label{tab:cm}Confusion matrix of the labels assigned by annotators to each instance with aggregated results for all languages. 
    For each column, the highlighted value corresponds to cases in which both annotators agreed on the annotation label. Subsequent values in each column correspond to cases in which one annotator chose the label corresponding to the column and the other one chose a different label.}
    \vspace{-0.2cm}
\end{table}

\vspace{-0.1cm}
\paragraph{Full results} The full results of the annotation are shown in Table~\ref{tab:human_mono}. We distinguish between ``Running Text, but slightly non-standard'' (RT) and ``Publishable Text'' (PT) in our annotation scheme but, for the purpose of training LMs, we consider both categories appropriate. In fact, language models might actually benefit from also observing non-standard language use during training.
Therefore, we also show the aggregated score of these two categories. Similarly, the other three categories are considered problematic for language model training. One exception is the category ``Wrong Language'' for Serbian, Croatian, Bosnian and Montenegrin: annotators were asked to distinguish between them, but a paragraph in Montenegrin instead of Serbian is very likely to be considered useful when training a Serbian LM. Actually, in the next section we only train a single language model for Croatian and Serbian. Generally speaking, we observe that MaCoCu and OSCAR contain paragraphs that are most often at least running text. MaCoCu has the highest score for 5 out of the 9 languages for which a comparison can be made, while OSCAR has the highest quality corpus for the other 4 languages. MC4 is the most problematic corpus: it has the least amount of useful paragraphs for all 8 languages it was included in. Especially Maltese seems to have issues: 164 out of 200 instances were in the wrong language for mC4, while half the MaCoCu instances did not contain running text.

\begin{table}[!t]
    \centering
    \setlength{\tabcolsep}{4pt}
    \resizebox{\columnwidth}{!}{
    \begin{tabular}{lccccc|c}
         \toprule
        &    \textbf{WL} & \textbf{NR} & \textbf{PR} & \textbf{RT} & \textbf{PT}  & \textbf{RT+PT} \\
      \midrule
      \textbf{MaCoCu}  & 1.8 &   3.6 &  10.4 &  29.9 & 54.3 &    \tgreen{84.2}     \\
      \textbf{CC100}    & 0.4 &   6.9 &  13.3 &  26.2 & 53.2 &    79.4     \\
        \textbf{mC4}      &  6.4 &  13.9 &  16.0 &  22.4 & 41.3 &    \tred{63.7}      \\
         \textbf{OSCAR}  &   0.6 &   5.4 &   9.9 &  24.7 & 59.4 &    84.1   \\
         \bottomrule
    \end{tabular}
    }
    \vspace{-0.1cm}
    \caption{\label{tab:avg_human_mono}Percentage of annotations for each of the annotation categories, averaged over corpus across the \textbf{seven} languages included in all evaluated corpora.}
    \vspace{-0.1cm}
\end{table}

\vspace{-0.1cm}
\paragraph{Average scores} To get a clearer overview of the quality of each corpus, we also show an averaged score of the corpora involved. For a fair comparison, we only average over the seven languages (Albanian, Bulgarian, Icelandic, Macedonian, Serbian, Slovenian and Turkish) included in all of the four evaluated corpora. We do not show the total counts but the percentage of each annotation and average across the seven languages. This is shown in Table~\ref{tab:avg_human_mono}. In this scenario, MaCoCu and OSCAR still seem to be the highest quality corpora, with CC100 not far behind. The mC4 corpus is clearly of lower quality than the other three.
Generally speaking, the results of this annotation do paint a slightly worrying picture about web-crawled monolingual data. For example, for mC4, around 1 out of every 5 paragraphs has serious issues: being in the wrong language or not (completely) consisting of running text. What might be even worse is that, for all corpora, only around half the paragraphs are of publishable quality, while the standards for this category were not particularly strict.

\begin{figure}[!tb]
 \begin{center}
 \hspace{-2mm} \includegraphics[width=0.485\textwidth]{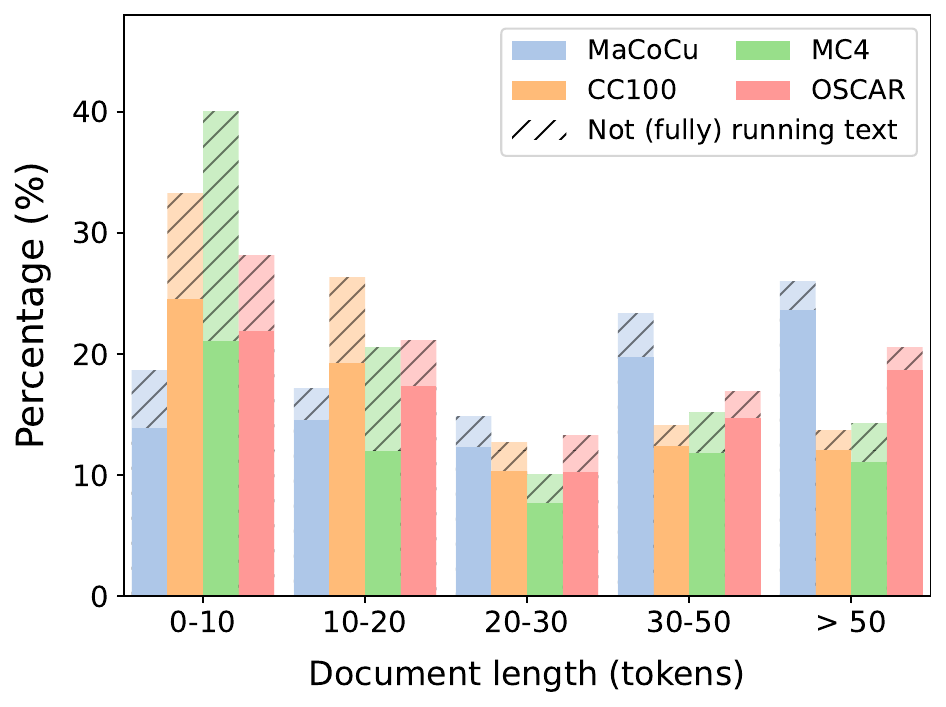}
 \vspace{-0.4cm}
\caption{Percentage of annotated documents that are of certain length for each corpus, averaged over the seven languages that had data available in each corpus. For each bar we indicate the percentage of documents that did not fully contain running text, i.e., were annotated as Wrong Language, Not-running Text or Partially Running Text.}
\label{fig:doc_length}
\end{center}
\vspace{-0.2cm}
\end{figure}

\vspace{-0.1cm}
\paragraph{Document length} We now examine the impact of the varying document lengths within each corpus. It can be argued that a corpus with numerous lengthy documents might still offer greater utility for language model training, even if it contains a lower percentage of high-quality documents. Moreover, longer documents are perhaps more likely to not be of publishable quality (since there is simply more text that can have issues), while still potentially preferable over very short, but high-quality documents. In Figure~\ref{fig:doc_length}, we plot the percentage of documents that fit into a certain document length range for each corpus. For each value, we also indicate what percentage was annotated as not fully running text (i.e., WL, NR or PR). The percentages shown here are averaged over all seven languages that had data for each of the four corpora. We can clearly see that the two corpora that had the highest amount of running and publishable text (MaCoCu and OSCAR) also have the highest percentage of large documents. Therefore, we are confident that the two potential issues identified above did not unfairly influence our annotation results.

\section{Automatic Evaluation}

This section focuses on the automatic evaluation of the corpora. We evaluate the corpora 
extrinsically by training general purpose encoder-only LMs. 
As our budget is limited, we evaluate on a subset of the languages in the manual evaluation, including: Albanian, Croatian, Icelandic, Serbian and Slovenian.

\vspace{-0.1cm}
\paragraph{Continued training \& Data} We do not start training models from scratch, but continue training XLM-R \citep{xlmr} for each corpus and language, as this is a more realistic usage of the often limited available data, and it is more computationally efficient. We opt for the base model instead of the large variant as the performance differences were small in initial experiments, while running the large model is approximately twice as expensive.
We also train a system per language that concatenates all four available corpora as a comparison. 
Data sizes per corpus and language are shown in Table~\ref{tab:size_lm}.
We train only a single model for Croatian and Serbian, as the languages are very similar and mutually intelligible, though we do transliterate all Cyrillic Serbian and Croatian data to Latin.\footnote{The Croatian corpora had < 0.1\% of Cyrillic data.} Note that the mC4 corpus does not treat Croatian as a separate language, while OSCAR  has very little Croatian data. 

\begin{table}[!t]
    \centering
    \setlength{\tabcolsep}{3pt}
    \resizebox{\columnwidth}{!}{
    \begin{tabular}{lrrrrr}
         \toprule
         \textbf{Language} & \multicolumn{1}{c}{\textbf{CC100}} & \multicolumn{1}{c}{\textbf{MaCoCu}} & \multicolumn{1}{c}{\textbf{mC4}} & \multicolumn{1}{c}{\textbf{OSCAR}} & \multicolumn{1}{c}{\textbf{Comb}} \\
        \midrule
        Albanian  & 2.1   & 1.4   & 4.6    & 0.9 &  9.0  \\
        Icelandic & 1.3  & 1.6   & 2.9   & 0.6 & 6.3   \\
        Serbo-Croatian   & 10.8   & 12.1   & 4.9   & 1.4 &  29.2  \\
        Slovenian   & 4.2   & 4.7    & 11.0  & 0.4 &  20.1  \\
        \midrule
        Croatian & 8.6 & 5.9  & ---  & 0.003 &  --- \\
        Serbian  & 2.2 & 6.2  & 4.9  & 1.4  & ---\\
         \bottomrule
    \end{tabular}
    }
    \vspace{-0.1cm}
    \caption{\label{tab:size_lm}Data set sizes in GB of compressed text for the included corpora in LM experiments. Serbian and Croatian individual figures are included for reference although only a single model is trained.}
    \vspace{-0.1cm}
\end{table}

\vspace{-0.1cm}
\paragraph{Details}
As previously stated, we continue training the XLM-R-base model. Each model is trained for 50,000 steps. We use a batch size of 1,024, a max learning rate of 1e-4 and 5,000 steps as warm-up. We do not make any modifications to the vocabulary.
In total, we train 20 different LMs, one for each corpus-language combination (with Serbo-Croatian being treated as single language). A single experiment (50,000 steps) took around 4 days on a single Google Cloud TPU.

\vspace{-0.1cm}
\paragraph{Fine-tuning} The trained LMs are evaluated by fine-tuning them on downstream tasks. Even though we trained a single model for Serbo-Croatian, we evaluate Serbian and Croatian separately. We use the same tasks across all five languages: language-specific Part-of-Speech tagging (XPOS), Named Entity Recognition (NER), Choice of Plausible Alternatives (COPA, \citealp{copa}) and Commitment Bank (CB, \citealp{cb}). The first two are classic evaluation tasks in NLP, while the latter two are part of the well-known SuperGLUE benchmark for English \citep{wang2019superglue}. 
For XPOS and NER, we use data from the Universal Dependencies project\footnote{\url{https://universaldependencies.org/}}, with the exception of Icelandic NER, for which we use the MIM-GOLD-NER set~\cite{mim-gold-ner}.
The COPA data set was originally created for just English \cite{copa}, but has gold standard translations available for Croatian, Serbian and Slovenian \citeplanguageresource{COPA-HR,COPA-SR,COPA-SI}. For Albanian and Icelandic, we translated the English data with Google Translate. For the CB task, we used Google Translate for all languages.\footnote{All translations were obtained in June 2023.} We use the suggested train, development and test splits for each task. For each language-task combination, we tune the learning rate of XLM-R-base on the development set and subsequently use this learning rate across other experiments. For XPOS/NER we report weighted F-score, while for COPA and CB we report accuracy. Further details regarding data set sizes, training regimes and hyper-parameter settings are described in the Appendix.

\subsection{Results}

First, we evaluate performance after training for 50,000 steps. We average each of these evaluations over 10, 20 and 30 different random seeds for XPOS/NER, COPA and CB, respectively.
For each language, we also report the \emph{position} in the ranking of each corpus. This gives a clearer overview of the performance of the corpus despite the fact that such a ranking does not show the relative differences in the scores. Two models are considered to have a different position if they differ significantly according to the Mann-Whitney test \cite{mann-whitney}. Note that a lower number for position means that the model performed better.

\begin{table}[!htb]
    \centering
        \setlength{\tabcolsep}{2pt}
    \resizebox{\columnwidth}{!}{
    \begin{tabular}{l|r|cccc|cccc|c}
         \toprule
        & & \multicolumn{4}{c|}{\textbf{Scores}} & \multicolumn{4}{c}{\textbf{Positions}} \\ 
       \multicolumn{1}{c|}{\textbf{Corpus}}  &  \multicolumn{1}{c|}{\textbf{Ep.}} &  \textbf{NER} & \textbf{POS} & \textbf{CP} & \textbf{CB} &  \textbf{NER} & \textbf{POS} & \textbf{CP} & \textbf{CB} & \textbf{Avg.} \\ 
         \midrule
      \textbf{Croatian} & & & & & & & & & \\
         \tabspace XLM-R & --- & 89.4 & 93.4 & 58.4 & 77.8 & 5   & 6   & 1    & 5  & 4.25 \\
       \tabspace CC100 & 3.6 & 90.9 & 94.4 & 59.8 & 80.1 & 1   & 1   & 1    & 1  & 1.00  \\
        \tabspace MaCoCu & 3.2 & 90.7 & 94.3 & 59.2 & 77.9 & 1   & 1   & 1    & 5  & 2.00  \\
        \tabspace mC4 & 7.4 & 89.8 & 94.2 & 55.8 & 80.1 & 4   & 4   & 5    & 1  & 3.50 \\
        \tabspace OSCAR & 24.9 & 89.3 & 94.1 & 56.2 & 78.9 & 5   & 5   & 5    & 1  & 4.00\\
       \tabspace Combined & 1.3 & 90.8 & 94.3 & 58.6 & 79.6 & 1   & 1   & 1    & 1  & 1.00\\
        \midrule
        \textbf{Serbian} & & & & & & & & & \\
        \tabspace XLM-R & ---  & 93.5 & 91.0 & 57.6 & 76.6 & 5   & 6   & 3    & 5  & 4.75\\
       \tabspace CC100 & 3.6  & 94.6 & 92.6 & 61.2 & 80.3 & 1   & 1   & 1    & 1  & 1.00\\
        \tabspace MaCoCu & 3.2 & 94.5 & 92.4 & 58.3 & 76.2 & 1   & 3   & 3    & 5  & 3.00\\
        \tabspace mC4 & 7.4 & 93.8 & 92.4 & 53.8 & 80.9 & 4   & 3   & 6    & 1  & 3.50 \\
        \tabspace OSCAR & 24.9  & 93.5 & 92.4 & 58.0 & 77.9 & 5   & 3   & 3    & 3  & 3.50 \\
       \tabspace Combined & 1.3& 94.6 & 92.4 & 60.4 & 78.3 & 1   & 1   & 1    & 3  & 1.50  \\
        \midrule
        \textbf{Albanian} & & & & & & & & & \\
        \tabspace XLM-R& --- & 92.7 & 93.9 & 54.9 & 77.5 & 3   & 5   & 6    & 1  & 3.75\\
       \tabspace CC100 & 17.2  & 92.9 & 94.1 & 60.8 & 78.2 & 3   & 1   & 1    & 1  & 1.50  \\
        \tabspace MaCoCu & 26.2 & 93.2 & 94.0 & 57.8 & 76.5 & 1   & 2   & 3    & 5  & 2.75 \\
        \tabspace mC4 & 7.8 & 92.8 & 94.0 & 56.4 & 77.8 & 3   & 2   & 4    & 1  & 2.50 \\
        \tabspace OSCAR & 37.2 & 92.8 & 94.0 & 55.7 & 79.5 & 3   & 2   & 4    & 1  & 2.50\\
       \tabspace Combined & 4.2 & 93.1 & 93.9 & 59.7 & 76.6 & 1   & 5   & 2    & 5  & 3.25\\
        \midrule
      \textbf{Icelandic} & & & & & & & & & \\
        \tabspace XLM-R & --- & 83.9 & 92.0 & 54.6 & 75.1 & 6   & 6   & 6    & 1  & 4.75 \\
       \tabspace CC100 & 28.1 & 88.1 & 93.6 & 59.1 & 74.8 & 1   & 4   & 1    & 1  & 1.75 \\
        \tabspace MaCoCu & 23.1 & 88.2 & 93.8 & 58.5 & 73.9 & 1   & 1   & 1    & 1  & 1.00  \\
        \tabspace mC4 & 11.5 & 87.8 & 93.8 & 55.8 & 75.1 & 1   & 1   & 4    & 1  & 1.75\\
        \tabspace OSCAR & 53.1 & 87.6 & 93.5 & 59.4 & 74.1 & 4   & 4   & 1    & 1  & 2.50 \\
       \tabspace Combined & 5.6 & 88.1 & 93.7 & 58.2 & 74.6 & 1   & 1   & 1    & 6  & 2.25\\
        \midrule
        \textbf{Slovenian} & & & & & & & & & \\
         \tabspace XLM-R & --- & 88.8 & 94.0 & 54.7 & 77.0 & 6   & 6   & 4    & 1  & 4.25 \\
       \tabspace CC100 & 8.8 & 90.7 & 95.6 & 56.6 & 76.0 & 1   & 1   & 1    & 1  & 1.00\\
        \tabspace MaCoCu & 6.6  & 90.1 & 95.3 & 53.9 & 76.0 & 4   & 4   & 4    & 1  & 3.25\\
        \tabspace mC4 & 3.3 & 90.4 & 95.6 & 54.5 & 77.3 & 1   & 1   & 4    & 1  & 1.75 \\
        \tabspace OSCAR & 96.3 & 89.8 & 95.4 & 56.8 & 76.5 & 4   & 4   & 1    & 1  & 2.50 \\
       \tabspace Combined & 1.8  & 90.7 & 95.6 & 56.7 & 75.6 & 1   & 1   & 1    & 6  & 2.25 \\
         \bottomrule
    \end{tabular}
    }
    \vspace{0.2cm}
    \caption{\label{tab:lm_res}Evaluation results for our 5 languages for XPOS, NER, COPA (CP) and CB. For Albanian, only UPOS data was available. Reported scores are averaged over 10, 20 and 30 runs for POS/NER, COPA and CB, respectively. Ep. denotes the number of epochs 
    corresponding to 50,000 steps. We consider a position to be different if we find a significant difference between two systems when using the Mann-Whitney test \citep{mann-whitney}.}
    \vspace{0.6cm}
\end{table}

\para{Full results} All results are shown in Table~\ref{tab:lm_res}. Since XPOS is a relatively easy task, the differences between the corpora here are, as expected, quite small. For the other tasks, there is a bit more variation. One thing that stands out is that we improve on the XLM-R baseline in virtually all settings. Continuing training a multi-lingual language model on a specific language of interest is a simple and (relatively) cheap way of improving performance. There is also quite some variance in the results. For example, the Serbo-Croatian model trained on mC4 obtains the best performance on CB for both languages, while getting the worst performance on the COPA task (even worse than the baseline). 

\para{Averaged results} Nevertheless, to obtain a clearer overview of performance per corpus, we show the averaged relative rankings in Table~\ref{tab:avg_pos}. 
We observe a surprising result: the best performing model seems to be based on the CC100 corpus, even ahead of the combined model. Though differences are small, the OSCAR corpus seems to be the worst corpus for LM training. This can likely be attributed to the fact that it is generally the smallest (see Table~\ref{tab:size_lm}). 

\begin{table}[!tb]
    \centering
        \setlength{\tabcolsep}{5pt}
    \resizebox{\columnwidth}{!}{
    \begin{tabular}{l|ccccc|c}
         \toprule
       &   \textbf{hr} & \textbf{sr} & \textbf{sq} & \textbf{is} & \textbf{sl} & \textbf{Avg. }\\ 
   \midrule
     XLM-R   & 4.25 & 4.75 & 3.75 & 4.75 & 4.25 & 4.35 \\
    CC100  & 1.0  & 1.0  & 1.5  & 1.75 & 1.0  & 1.25 \\
     MaCoCu & 2.0  & 3.0  & 2.75 & 1.0  & 3.25 & 2.40  \\
     mC4    & 3.5  & 3.5  & 2.5  & 2.75 & 1.75 & 2.80  \\
     OSCAR  & 4.0  & 3.5  & 2.5  & 2.75 & 2.5  & 3.05 \\
     Comb   & 1.0  & 1.5  & 3.25 & 1.75 & 2.25 & 1.95 \\
         \bottomrule
    \end{tabular}
    }
    \vspace{0.1cm}
    \caption{\label{tab:avg_pos}Results for each corpus when averaging the position for each language (i.e., over 4 tasks, see Table~\ref{tab:lm_res}), and finally (last column) averaging over all the languages. Each model was trained for \textbf{50,000 steps}.}
\end{table}

\begin{figure}[!tb]
 \begin{center}
 \vspace{-0.75cm}
 \includegraphics[width=0.53\textwidth]{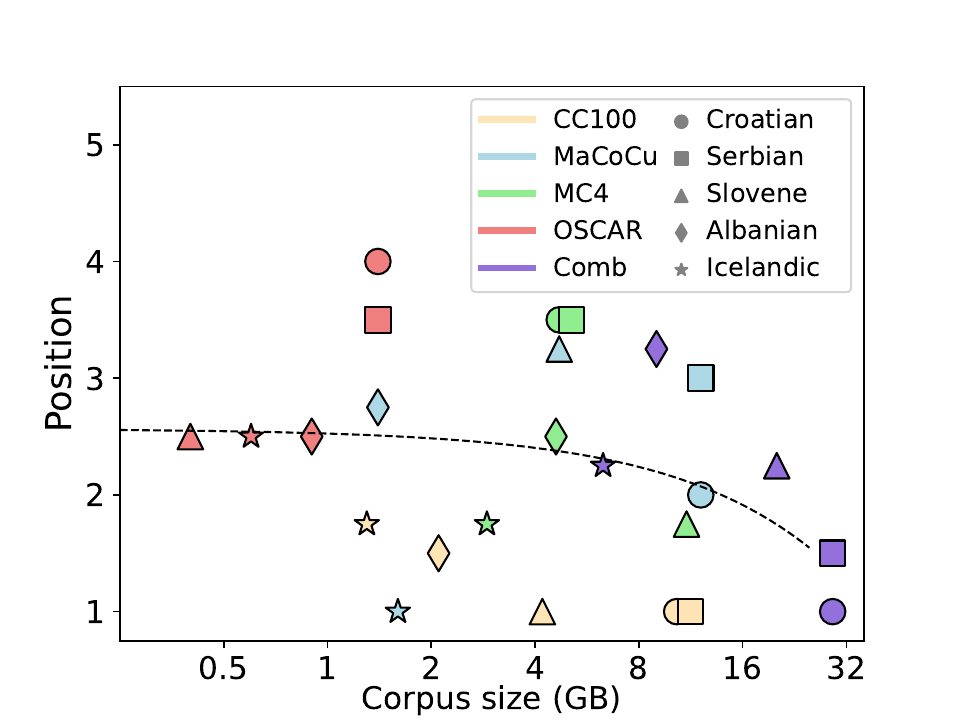}
\caption{Average position across the four evaluation tasks plotted over the data set size (GB), for each language-corpus combination. Dotted line is the linear regression line. Note the log scale of the X-axis.}
\label{fig:scatter}
\end{center}
\end{figure}

\para{Importance of data set size} So far, we explicitly did not control for data set size. The size of the released data is in many ways a design choice: stricter filtering leads to higher quality data, but this might not be preferable when training data-hungry LMs. It is therefore not entirely fair on a given corpus to run experiments in which data set size is controlled for. Nevertheless, we want to investigate how much the results are influenced simply by the amount of unique data available for each corpus, irrespective of data quality. Therefore, we plot the average position (where lower position means better performance) over the size of the data set used for each corpus in Figure~\ref{fig:scatter}.
As expected,
models trained on smaller corpora tend to underperform, though there only seems to be a small effect. However, note that there is a clear confounding factor here: the only corpus that is noticeably smaller for all languages is OSCAR, making it unclear whether the relatively bad performance is due to the corpus or the amount of unique data.

\para{Controlling for size} Therefore, we recreate the results in Table~\ref{tab:lm_res} and \ref{tab:avg_pos}, but now for the results after just 10,000 steps instead of 50,000. At this stage, the models did not see much data yet, so the effect of size should not play a (large) role here, as was also previously shown in \citet{muennighoff2023scaling}. For space reasons, we only show the averaged results in Table~\ref{tab:avg_pos_controlled}. The CC100 and combined corpus still have the best performance, while the other three corpora are similar.
Surprisingly, even in the data-controlled setting, the perceived quality of the data by humans does not seem to influence the downstream performance of LMs. Nevertheless, the performance of OSCAR is now similar to that of mC4 and MaCoCu, suggesting that it indeed was disadvantaged by its size in the previous experiments.

\paragraph{XLM-R pretraining corpus} The CC100 corpus was in fact already used for pretraining XLM-R. This could potentially be a disadvantage for this corpus, as the data will never be completely new when continuing to train XLM-R. However, given the fact that our languages only make up a very small part of the full corpus\footnote{Croatian is the language with the most data, occurring in the 30th position of the 100 languages involved.}, we did not expect that this would make a large difference. In fact, given the excellent performance of CC100, we can be reasonably sure that this did not negatively affect the results of the CC100 corpus. On the contrary: it is looking more like it \emph{positively} affected the results of CC100, especially since it also outperformed the combined corpus. We leave investigating why and how this could be the case for future work.

\begin{table}[!tb]
    \centering
        \setlength{\tabcolsep}{5pt}
    \resizebox{\columnwidth}{!}{
    \begin{tabular}{l|ccccc|c}
         \toprule
       &   \textbf{hr} & \textbf{sr} & \textbf{sq} & \textbf{is} & \textbf{sl} & \textbf{Avg. }\\ 
   \midrule
     XLM-R   & 3.5  & 3.5 & 3.5  & 4.75 & 4.0  & 3.85 \\
    CC100  &1.0  & 1.0 & 1.5  & 2.5  & 1.0  & 1.40  \\
     MaCoCu &  1.5  & 2.5 & 2.0  & 2.75 & 4.25 & 2.60  \\

     mC4    & 2.25 & 2.5 & 2.25 & 3.0  & 2.5  & 2.50  \\
     OSCAR  & 3.75 & 2.5 & 2.0  & 3.0  & 2.25 & 2.70  \\
     Comb   & 1.0  & 1.0 & 2.25 & 1.5  & 2.25 & 1.60 \\
         \bottomrule
    \end{tabular}
    }
    \vspace{0.2cm}
    \caption{\label{tab:avg_pos_controlled}Results for each corpus when averaging the position for each language (i.e., over 4 tasks, see Table~\ref{tab:lm_res}), and finally (last column) averaging over all the languages. Each model was trained for \textbf{10,000 steps}.}
    \vspace{0.2cm}
\end{table}

\vspace{0.1cm}
\section{Conclusion}
\vspace{0.1cm}

Even though large, curated, web-crawled corpora form the lion's share of the training data of all popular language models, the quality of these corpora has been given relatively little attention. In this paper, we compared four of the currently most relevant large, web-crawled corpora (CC100, MaCoCu, mC4 and OSCAR) across eleven lower-resourced European languages. We first performed a human evaluation by hiring professional linguists to annotate the (rudimentary) quality of the corpora. We found clear differences between the corpora: MaCoCu and OSCAR were of higher quality than CC100 and mC4. We then performed an automatic evaluation of the corpora on a subset of five languages by training language models on each language-corpus combination and evaluation performance on downstream applications. Surprisingly, we found that CC100 is the corpus that has the best performance, even if we control for data set size. We therefore conclude that data set quality (as judged by humans) of web-crawled corpora does not seem to play a significant role in training language models. 

\section{Limitations}
Clearly, our work has a number of limitations. For one, our annotation scheme is quite rudimentary: annotators mainly have to determine whether a given paragraph consists of running text or not. However, we believe this is the most vital characteristic of the text when training language models. Moreover, the use of a more complex annotation scheme has major challenges in making sure the results are comparable across different annotators and languages. Secondly, we train encoder-only language models and not generative ones such as GPT4 or LLAMA. It is conceivable that data set quality plays a larger role when models actually have to generate text. We plan to investigate this in future work. Thirdly, we only evaluate on European languages. This was driven by the fact that we wanted to compare multiple corpora, with MaCoCu only working with European languages. Since we do have quite some variety within our languages, we are confident that our results still generalize. Lastly, our models are evaluated on just four tasks, only two of which concern natural language understanding. The main limitation here is simply availability: there are not many evaluation tasks available across a large number of languages. Since we wanted to compare corpora across languages, we opted to only include tasks that had data available for all languages used in our study.

\section{Acknowledgements}

The MaCoCu project has received funding from the European Union’s Connecting Europe Facility 2014-2020 - CEF Telecom, under Grant Agreement No. INEA/CEF/ICT/A2020/2278341. This communication reflects only the authors’ views. The Agency is not responsible for any use that may be made of the information it contains. This work was also funded by the Slovenian Research Agency within the national projects J7-4642 and L2-50070, as well the research programme P6-0411. This research was supported with Cloud TPUs from Google's TPU Research Cloud (TRC). We thank the Center for Information Technology of the University of Groningen for providing access to the Hábrók high performance computing cluster. We are grateful to all our MaCoCu colleagues for the fruitful discussions. We thank Wietse de Vries for his feedback on the annotation scheme and for agreeing to do a pilot annotation run. Finally, we thank the reviewers for their comments.

\newpage
\section{Bibliographical References}\label{reference}

\bibliographystyle{lrec-coling2024-natbib}
\bibliography{lrec-coling2024-example}

\begin{thebibliography}{3}
\expandafter\ifx\csname natexlab\endcsname\relax\def\natexlab#1{#1}\fi

\bibitem[{Ljube{\v s}i{\'c}(2021)}]{COPA-HR}
Ljube{\v s}i{\'c}, Nikola. 2021.
\newblock \href {http://hdl.handle.net/11356/1404} {\emph{Choice of plausible
  alternatives dataset in Croatian {COPA}-{HR}}}.
\newblock Slovenian language resource repository {CLARIN}.{SI}.

\bibitem[{Ljube{\v s}i{\'c} et~al.(2022)Ljube{\v s}i{\'c}, Starovi{\'c},
  Kuzman, and Samard{\v z}i{\'c}}]{COPA-SR}
Ljube{\v s}i{\'c}, Nikola and Starovi{\'c}, Mirjana and Kuzman, Taja and
  Samard{\v z}i{\'c}, Tanja. 2022.
\newblock \href {http://hdl.handle.net/11356/1708} {\emph{Choice of plausible
  alternatives dataset in Serbian {COPA}-{SR}}}.
\newblock Slovenian language resource repository {CLARIN}.{SI}.

\bibitem[{{\v Z}agar et~al.(2020){\v Z}agar, Robnik-{\v S}ikonja, Goli, and
  Arhar~Holdt}]{COPA-SI}
{\v Z}agar, Ale{\v s} and Robnik-{\v S}ikonja, Marko and Goli, Teja and Arhar
  Holdt, {\v S}pela. 2020.
\newblock \href {http://hdl.handle.net/11356/1380} {\emph{Slovene translation
  of {SuperGLUE}}}.
\newblock Slovenian language resource repository {CLARIN}.{SI}.

\end{thebibliography}


\begin{thebibliography}{45}
\expandafter\ifx\csname natexlab\endcsname\relax\def\natexlab#1{#1}\fi

\bibitem[{Artetxe et~al.(2022)Artetxe, Aldabe, Agerri, Perez-de Vi{\~n}aspre,
  and Soroa}]{artetxe-etal-2022-corpus}
Mikel Artetxe, Itziar Aldabe, Rodrigo Agerri, Olatz Perez-de Vi{\~n}aspre, and
  Aitor Soroa. 2022.
\newblock \href {https://aclanthology.org/2022.emnlp-main.499} {Does corpus
  quality really matter for low-resource languages?}
\newblock In \emph{Proceedings of the 2022 Conference on Empirical Methods in
  Natural Language Processing}, pages 7383--7390, Abu Dhabi, United Arab
  Emirates. Association for Computational Linguistics.

\bibitem[{Ba{\~n}{\'o}n et~al.(2022)Ba{\~n}{\'o}n, Espl{\`a}-Gomis, Forcada,
  Garc{\'\i}a-Romero, Kuzman, Ljube{\v{s}}i{\'c}, van Noord, Sempere,
  Ram{\'\i}rez-S{\'a}nchez, Rupnik, Suchomel, Toral, van~der Werff, and
  Zaragoza}]{macocu}
Marta Ba{\~n}{\'o}n, Miquel Espl{\`a}-Gomis, Mikel~L. Forcada, Cristian
  Garc{\'\i}a-Romero, Taja Kuzman, Nikola Ljube{\v{s}}i{\'c}, Rik van Noord,
  Leopoldo~Pla Sempere, Gema Ram{\'\i}rez-S{\'a}nchez, Peter Rupnik, V{\'\i}t
  Suchomel, Antonio Toral, Tobias van~der Werff, and Jaume Zaragoza. 2022.
\newblock \href {https://aclanthology.org/2022.eamt-1.41} {{M}a{C}o{C}u:
  Massive collection and curation of monolingual and bilingual data: focus on
  under-resourced languages}.
\newblock In \emph{Proceedings of the 23rd Annual Conference of the European
  Association for Machine Translation}, pages 303--304, Ghent, Belgium.
  European Association for Machine Translation.

\bibitem[{Bender et~al.(2021)Bender, Gebru, McMillan-Major, and
  Shmitchell}]{bender2021}
Emily~M. Bender, Timnit Gebru, Angelina McMillan-Major, and Shmargaret
  Shmitchell. 2021.
\newblock \href {https://doi.org/10.1145/3442188.3445922} {On the dangers of
  stochastic parrots: Can language models be too big?}
\newblock In \emph{Proceedings of the 2021 ACM Conference on Fairness,
  Accountability, and Transparency}, FAccT '21, page 610–623, New York, NY,
  USA. Association for Computing Machinery.

\bibitem[{Caswell et~al.(2020)Caswell, Breiner, van Esch, and
  Bapna}]{caswell-etal-2020-language}
Isaac Caswell, Theresa Breiner, Daan van Esch, and Ankur Bapna. 2020.
\newblock \href {https://doi.org/10.18653/v1/2020.coling-main.579} {Language
  {ID} in the wild: Unexpected challenges on the path to a thousand-language
  web text corpus}.
\newblock In \emph{Proceedings of the 28th International Conference on
  Computational Linguistics}, pages 6588--6608, Barcelona, Spain (Online).
  International Committee on Computational Linguistics.

\bibitem[{Chau et~al.(2020)Chau, Lin, and Smith}]{chau2020parsing}
Ethan~C. Chau, Lucy~H. Lin, and Noah~A. Smith. 2020.
\newblock \href {https://doi.org/10.18653/v1/2020.findings-emnlp.118} {Parsing
  with multilingual {BERT}, a small corpus, and a small treebank}.
\newblock In \emph{Findings of the Association for Computational Linguistics:
  EMNLP 2020}, pages 1324--1334, Online. Association for Computational
  Linguistics.

\bibitem[{Conneau et~al.(2020)Conneau, Khandelwal, Goyal, Chaudhary, Wenzek,
  Guzm{\'a}n, Grave, Ott, Zettlemoyer, and Stoyanov}]{xlmr}
Alexis Conneau, Kartikay Khandelwal, Naman Goyal, Vishrav Chaudhary, Guillaume
  Wenzek, Francisco Guzm{\'a}n, Edouard Grave, Myle Ott, Luke Zettlemoyer, and
  Veselin Stoyanov. 2020.
\newblock \href {https://doi.org/10.18653/v1/2020.acl-main.747} {Unsupervised
  cross-lingual representation learning at scale}.
\newblock In \emph{Proceedings of the 58th Annual Meeting of the Association
  for Computational Linguistics}, pages 8440--8451, Online. Association for
  Computational Linguistics.

\bibitem[{De~Marneffe et~al.(2019)De~Marneffe, Simons, and Tonhauser}]{cb}
Marie-Catherine De~Marneffe, Mandy Simons, and Judith Tonhauser. 2019.
\newblock The commitmentbank: Investigating projection in naturally occurring
  discourse.
\newblock In \emph{proceedings of Sinn und Bedeutung}, volume~23, pages
  107--124.

\bibitem[{de~Vries et~al.(2019)de~Vries, van Cranenburgh, Bisazza, Caselli, van
  Noord, and Nissim}]{bertje}
Wietse de~Vries, Andreas van Cranenburgh, Arianna Bisazza, Tommaso Caselli,
  Gertjan van Noord, and Malvina Nissim. 2019.
\newblock Bertje: A dutch bert model.
\newblock \emph{arXiv preprint arXiv:1912.09582}.

\bibitem[{Dodge et~al.(2021)Dodge, Sap, Marasovi{\'c}, Agnew, Ilharco,
  Groeneveld, Mitchell, and Gardner}]{dodge-etal-2021-documenting}
Jesse Dodge, Maarten Sap, Ana Marasovi{\'c}, William Agnew, Gabriel Ilharco,
  Dirk Groeneveld, Margaret Mitchell, and Matt Gardner. 2021.
\newblock \href {https://doi.org/10.18653/v1/2021.emnlp-main.98} {Documenting
  large webtext corpora: A case study on the colossal clean crawled corpus}.
\newblock In \emph{Proceedings of the 2021 Conference on Empirical Methods in
  Natural Language Processing}, pages 1286--1305, Online and Punta Cana,
  Dominican Republic. Association for Computational Linguistics.

\bibitem[{Ebrahimi and Kann(2021)}]{ebrahimi2021adapt}
Abteen Ebrahimi and Katharina Kann. 2021.
\newblock \href {https://doi.org/10.18653/v1/2021.acl-long.351} {How to adapt
  your pretrained multilingual model to 1600 languages}.
\newblock In \emph{Proceedings of the 59th Annual Meeting of the Association
  for Computational Linguistics and the 11th International Joint Conference on
  Natural Language Processing (Volume 1: Long Papers)}, pages 4555--4567,
  Online. Association for Computational Linguistics.

\bibitem[{Grave et~al.(2018)Grave, Bojanowski, Gupta, Joulin, and
  Mikolov}]{grave-etal-2018-learning}
Edouard Grave, Piotr Bojanowski, Prakhar Gupta, Armand Joulin, and Tomas
  Mikolov. 2018.
\newblock \href {https://aclanthology.org/L18-1550} {Learning word vectors for
  157 languages}.
\newblock In \emph{Proceedings of the Eleventh International Conference on
  Language Resources and Evaluation ({LREC} 2018)}, Miyazaki, Japan. European
  Language Resources Association (ELRA).

\bibitem[{Gururangan et~al.(2020)Gururangan, Marasovi{\'c}, Swayamdipta, Lo,
  Beltagy, Downey, and Smith}]{gururangan-etal-2020-dont}
Suchin Gururangan, Ana Marasovi{\'c}, Swabha Swayamdipta, Kyle Lo, Iz~Beltagy,
  Doug Downey, and Noah~A. Smith. 2020.
\newblock \href {https://doi.org/10.18653/v1/2020.acl-main.740} {Don{'}t stop
  pretraining: Adapt language models to domains and tasks}.
\newblock In \emph{Proceedings of the 58th Annual Meeting of the Association
  for Computational Linguistics}, pages 8342--8360, Online. Association for
  Computational Linguistics.

\bibitem[{Heafield(2011)}]{heafield-2011-kenlm}
Kenneth Heafield. 2011.
\newblock \href {https://aclanthology.org/W11-2123} {{K}en{LM}: Faster and
  smaller language model queries}.
\newblock In \emph{Proceedings of the Sixth Workshop on Statistical Machine
  Translation}, pages 187--197, Edinburgh, Scotland. Association for
  Computational Linguistics.

\bibitem[{Ing{\'o}lfsd{\'o}ttir et~al.(2020)Ing{\'o}lfsd{\'o}ttir,
  Gu{dh}j{\'o}nsson, and Loftsson}]{mim-gold-ner}
Svanhv{\'{\i}}t~Lilja Ing{\'o}lfsd{\'o}ttir, {\'A}smundur~Alma
  Gu{dh}j{\'o}nsson, and Hrafn Loftsson. 2020.
\newblock \href {http://hdl.handle.net/20.500.12537/140} {{MIM}-{GOLD}-{NER}
  – named entity recognition corpus (21.09)}.
\newblock {CLARIN}-{IS}.

\bibitem[{Junczys-Dowmunt(2019)}]{junczys-dowmunt-2019-microsoft}
Marcin Junczys-Dowmunt. 2019.
\newblock \href {https://doi.org/10.18653/v1/W19-5321} {{M}icrosoft translator
  at {WMT} 2019: Towards large-scale document-level neural machine
  translation}.
\newblock In \emph{Proceedings of the Fourth Conference on Machine Translation
  (Volume 2: Shared Task Papers, Day 1)}, pages 225--233, Florence, Italy.
  Association for Computational Linguistics.

\bibitem[{Kreutzer et~al.(2022)Kreutzer, Caswell, Wang, Wahab, van Esch,
  Ulzii-Orshikh, Tapo, Subramani, Sokolov, Sikasote, Setyawan, Sarin, Samb,
  Sagot, Rivera, Rios, Papadimitriou, Osei, Suarez, Orife, Ogueji, Rubungo,
  Nguyen, M{\"u}ller, M{\"u}ller, Muhammad, Muhammad, Mnyakeni, Mirzakhalov,
  Matangira, Leong, Lawson, Kudugunta, Jernite, Jenny, Firat, Dossou, Dlamini,
  de~Silva, {\c{C}}abuk~Ball{\i}, Biderman, Battisti, Baruwa, Bapna, Baljekar,
  Azime, Awokoya, Ataman, Ahia, Ahia, Agrawal, and
  Adeyemi}]{kreutzer-etal-2022-quality}
Julia Kreutzer, Isaac Caswell, Lisa Wang, Ahsan Wahab, Daan van Esch,
  Nasanbayar Ulzii-Orshikh, Allahsera Tapo, Nishant Subramani, Artem Sokolov,
  Claytone Sikasote, Monang Setyawan, Supheakmungkol Sarin, Sokhar Samb,
  Beno{\^\i}t Sagot, Clara Rivera, Annette Rios, Isabel Papadimitriou, Salomey
  Osei, Pedro~Ortiz Suarez, Iroro Orife, Kelechi Ogueji, Andre~Niyongabo
  Rubungo, Toan~Q. Nguyen, Mathias M{\"u}ller, Andr{\'e} M{\"u}ller,
  Shamsuddeen~Hassan Muhammad, Nanda Muhammad, Ayanda Mnyakeni, Jamshidbek
  Mirzakhalov, Tapiwanashe Matangira, Colin Leong, Nze Lawson, Sneha Kudugunta,
  Yacine Jernite, Mathias Jenny, Orhan Firat, Bonaventure F.~P. Dossou, Sakhile
  Dlamini, Nisansa de~Silva, Sakine {\c{C}}abuk~Ball{\i}, Stella Biderman,
  Alessia Battisti, Ahmed Baruwa, Ankur Bapna, Pallavi Baljekar, Israel~Abebe
  Azime, Ayodele Awokoya, Duygu Ataman, Orevaoghene Ahia, Oghenefego Ahia,
  Sweta Agrawal, and Mofetoluwa Adeyemi. 2022.
\newblock \href {https://doi.org/10.1162/tacl_a_00447} {Quality at a glance: An
  audit of web-crawled multilingual datasets}.
\newblock \emph{Transactions of the Association for Computational Linguistics},
  10:50--72.

\bibitem[{Kuzman et~al.(2023)Kuzman, Mozeti{\v{c}}, and
  Ljube{\v{s}}i{\'c}}]{kuzman2023automatic}
Taja Kuzman, Igor Mozeti{\v{c}}, and Nikola Ljube{\v{s}}i{\'c}. 2023.
\newblock Automatic genre identification for robust enrichment of massive text
  collections: Investigation of classification methods in the era of large
  language models.
\newblock \emph{Machine Learning and Knowledge Extraction}, 5(3):1149--1175.

\bibitem[{Le et~al.(2020)Le, Vial, Frej, Segonne, Coavoux, Lecouteux, Allauzen,
  Crabb{\'e}, Besacier, and Schwab}]{flaubert}
Hang Le, Lo{\"\i}c Vial, Jibril Frej, Vincent Segonne, Maximin Coavoux,
  Benjamin Lecouteux, Alexandre Allauzen, Benoit Crabb{\'e}, Laurent Besacier,
  and Didier Schwab. 2020.
\newblock \href {https://aclanthology.org/2020.lrec-1.302} {{F}lau{BERT}:
  Unsupervised language model pre-training for {F}rench}.
\newblock In \emph{Proceedings of the Twelfth Language Resources and Evaluation
  Conference}, pages 2479--2490, Marseille, France. European Language Resources
  Association.

\bibitem[{Ljube{\v{s}}i{\'c} and Lauc(2021)}]{ljubesic-lauc-2021-bertic}
Nikola Ljube{\v{s}}i{\'c} and Davor Lauc. 2021.
\newblock \href {https://www.aclweb.org/anthology/2021.bsnlp-1.5} {{BERT}i{\'c}
  - the transformer language model for {B}osnian, {C}roatian, {M}ontenegrin and
  {S}erbian}.
\newblock In \emph{Proceedings of the 8th Workshop on Balto-Slavic Natural
  Language Processing}, pages 37--42, Kiyv, Ukraine. Association for
  Computational Linguistics.

\bibitem[{Luccioni and Viviano(2021)}]{luccioni-viviano-2021-whats}
Alexandra Luccioni and Joseph Viviano. 2021.
\newblock \href {https://doi.org/10.18653/v1/2021.acl-short.24} {What{'}s in
  the box? {A}n analysis of undesirable content in the {C}ommon {C}rawl
  corpus}.
\newblock In \emph{Proceedings of the 59th Annual Meeting of the Association
  for Computational Linguistics and the 11th International Joint Conference on
  Natural Language Processing (Volume 2: Short Papers)}, pages 182--189,
  Online. Association for Computational Linguistics.

\bibitem[{Mann and Whitney(1947)}]{mann-whitney}
H.~B. Mann and D.~R. Whitney. 1947.
\newblock \href {https://doi.org/10.1214/aoms/1177730491} {{On a Test of
  Whether one of Two Random Variables is Stochastically Larger than the
  Other}}.
\newblock \emph{The Annals of Mathematical Statistics}, 18(1):50 -- 60.

\bibitem[{Martin et~al.(2020)Martin, Muller, Ortiz~Su{\'a}rez, Dupont, Romary,
  de~la Clergerie, Seddah, and Sagot}]{camembert}
Louis Martin, Benjamin Muller, Pedro~Javier Ortiz~Su{\'a}rez, Yoann Dupont,
  Laurent Romary, {\'E}ric de~la Clergerie, Djam{\'e} Seddah, and Beno{\^\i}t
  Sagot. 2020.
\newblock \href {https://doi.org/10.18653/v1/2020.acl-main.645} {{C}amem{BERT}:
  a tasty {F}rench language model}.
\newblock In \emph{Proceedings of the 58th Annual Meeting of the Association
  for Computational Linguistics}, pages 7203--7219, Online. Association for
  Computational Linguistics.

\bibitem[{Muennighoff et~al.(2023)Muennighoff, Rush, Barak, Le~Scao, Tazi,
  Piktus, Pyysalo, Wolf, and Raffel}]{muennighoff2023scaling}
Niklas Muennighoff, Alexander Rush, Boaz Barak, Teven Le~Scao, Nouamane Tazi,
  Aleksandra Piktus, Sampo Pyysalo, Thomas Wolf, and Colin~A Raffel. 2023.
\newblock \href
  {https://proceedings.neurips.cc/paper_files/paper/2023/file/9d89448b63ce1e2e8dc7af72c984c196-Paper-Conference.pdf}
  {Scaling data-constrained language models}.
\newblock In \emph{Advances in Neural Information Processing Systems},
  volume~36, pages 50358--50376. Curran Associates, Inc.

\bibitem[{Muller et~al.(2021)Muller, Anastasopoulos, Sagot, and
  Seddah}]{muller2021being}
Benjamin Muller, Antonios Anastasopoulos, Beno{\^\i}t Sagot, and Djam{\'e}
  Seddah. 2021.
\newblock \href {https://doi.org/10.18653/v1/2021.naacl-main.38} {When being
  unseen from m{BERT} is just the beginning: Handling new languages with
  multilingual language models}.
\newblock In \emph{Proceedings of the 2021 Conference of the North American
  Chapter of the Association for Computational Linguistics: Human Language
  Technologies}, pages 448--462, Online. Association for Computational
  Linguistics.

\bibitem[{Oliver and Hagen(2021)}]{TLSH}
Jonathan Oliver and Josiah Hagen. 2021.
\newblock \href {https://doi.org/10.1109/EUC53437.2021.00028} {Designing the
  elements of a fuzzy hashing scheme}.
\newblock In \emph{2021 IEEE 19th International Conference on Embedded and
  Ubiquitous Computing (EUC)}, pages 1--6.

\bibitem[{OpenAI(2023)}]{GPT4}
OpenAI. 2023.
\newblock Gpt-4 technical report.
\newblock Technical report, OpenAI.

\bibitem[{{Ortiz Su{\'a}rez} et~al.(2019){Ortiz Su{\'a}rez}, Sagot, and
  Romary}]{oscar}
Pedro~Javier {Ortiz Su{\'a}rez}, Beno{\^i}t Sagot, and Laurent Romary. 2019.
\newblock \href {https://doi.org/10.14618/ids-pub-9021} {Asynchronous pipelines
  for processing huge corpora on medium to low resource infrastructures}.
\newblock Proceedings of the Workshop on Challenges in the Management of Large
  Corpora (CMLC-7) 2019. Cardiff, 22nd July 2019, pages 9 -- 16, Mannheim.
  Leibniz-Institut f{\"u}r Deutsche Sprache.

\bibitem[{Pfeiffer et~al.(2021)Pfeiffer, Kamath, R{\"u}ckl{\'e}, Cho, and
  Gurevych}]{pfeiffer2021adapterfusion}
Jonas Pfeiffer, Aishwarya Kamath, Andreas R{\"u}ckl{\'e}, Kyunghyun Cho, and
  Iryna Gurevych. 2021.
\newblock \href {https://doi.org/10.18653/v1/2021.eacl-main.39}
  {{A}dapter{F}usion: Non-destructive task composition for transfer learning}.
\newblock In \emph{Proceedings of the 16th Conference of the European Chapter
  of the Association for Computational Linguistics: Main Volume}, pages
  487--503, Online. Association for Computational Linguistics.

\bibitem[{Pfeiffer et~al.(2020)Pfeiffer, Vuli{\'c}, Gurevych, and
  Ruder}]{pfeiffer2020mad}
Jonas Pfeiffer, Ivan Vuli{\'c}, Iryna Gurevych, and Sebastian Ruder. 2020.
\newblock Mad-x: An adapter-based framework for multi-task cross-lingual
  transfer.
\newblock In \emph{Proceedings of the 2020 Conference on Empirical Methods in
  Natural Language Processing (EMNLP)}, pages 7654--7673.

\bibitem[{Rae et~al.(2021)Rae, Borgeaud, Cai, Millican, Hoffmann, Song,
  Aslanides, Henderson, Ring, Young et~al.}]{rae2021scaling}
Jack~W Rae, Sebastian Borgeaud, Trevor Cai, Katie Millican, Jordan Hoffmann,
  Francis Song, John Aslanides, Sarah Henderson, Roman Ring, Susannah Young,
  et~al. 2021.
\newblock Scaling language models: Methods, analysis \& insights from training
  gopher.
\newblock \emph{arXiv preprint arXiv:2112.11446}.

\bibitem[{Rajapakse(2019)}]{simpletransformers}
T.~C. Rajapakse. 2019.
\newblock Simple transformers.
\newblock \url{https://github.com/ThilinaRajapakse/simpletransformers}.

\bibitem[{Ranathunga and de~Silva(2022)}]{ranathunga-de-silva-2022-languages}
Surangika Ranathunga and Nisansa de~Silva. 2022.
\newblock \href {https://aclanthology.org/2022.aacl-main.62} {Some languages
  are more equal than others: Probing deeper into the linguistic disparity in
  the {NLP} world}.
\newblock In \emph{Proceedings of the 2nd Conference of the Asia-Pacific
  Chapter of the Association for Computational Linguistics and the 12th
  International Joint Conference on Natural Language Processing (Volume 1: Long
  Papers)}, pages 823--848, Online only. Association for Computational
  Linguistics.

\bibitem[{Roemmele et~al.(2011)Roemmele, Bejan, and Gordon}]{copa}
Melissa Roemmele, Cosmin~Adrian Bejan, and Andrew~S Gordon. 2011.
\newblock Choice of plausible alternatives: An evaluation of commonsense causal
  reasoning.
\newblock In \emph{AAAI spring symposium: logical formalizations of commonsense
  reasoning}, pages 90--95.

\bibitem[{Sanh et~al.(2019)Sanh, Debut, Chaumond, and
  Wolf}]{sanh2019distilbert}
Victor Sanh, Lysandre Debut, Julien Chaumond, and Thomas Wolf. 2019.
\newblock {DistilBERT}, a distilled version of {BERT}: smaller, faster, cheaper
  and lighter.
\newblock \emph{arXiv preprint arXiv:1910.01108}.

\bibitem[{Schweter(2020)}]{berturk}
Stefan Schweter. 2020.
\newblock \href {https://doi.org/10.5281/zenodo.3770924} {{BERTurk} - {BERT}
  models for {T}urkish}.

\bibitem[{Seker et~al.(2022)Seker, Bandel, Bareket, Brusilovsky, Greenfeld, and
  Tsarfaty}]{seker-etal-2022-alephbert}
Amit Seker, Elron Bandel, Dan Bareket, Idan Brusilovsky, Refael Greenfeld, and
  Reut Tsarfaty. 2022.
\newblock \href {https://doi.org/10.18653/v1/2022.acl-long.4} {{A}leph{BERT}:
  Language model pre-training and evaluation from sub-word to sentence level}.
\newblock In \emph{Proceedings of the 60th Annual Meeting of the Association
  for Computational Linguistics (Volume 1: Long Papers)}, pages 46--56, Dublin,
  Ireland. Association for Computational Linguistics.

\bibitem[{Sn{\ae}bjarnarson et~al.(2022)Sn{\ae}bjarnarson, S{\'\i}monarson,
  Ragnarsson, Ing{\'o}lfsd{\'o}ttir, J{\'o}nsson, Thorsteinsson, and
  Einarsson}]{icebert}
V{\'e}steinn Sn{\ae}bjarnarson, Haukur~Barri S{\'\i}monarson, P{\'e}tur~Orri
  Ragnarsson, Svanhv{\'\i}t~Lilja Ing{\'o}lfsd{\'o}ttir, Haukur J{\'o}nsson,
  Vilhjalmur Thorsteinsson, and Hafsteinn Einarsson. 2022.
\newblock \href {https://aclanthology.org/2022.lrec-1.464} {A warm start and a
  clean crawled corpus - a recipe for good language models}.
\newblock In \emph{Proceedings of the Thirteenth Language Resources and
  Evaluation Conference}, pages 4356--4366, Marseille, France. European
  Language Resources Association.

\bibitem[{Souza et~al.(2020)Souza, Nogueira, and Lotufo}]{souza2020bertimbau}
F{\'a}bio Souza, Rodrigo Nogueira, and Roberto Lotufo. 2020.
\newblock {BERTimbau: Pretrained BERT Models for Brazilian Portuguese}.
\newblock In \emph{Intelligent Systems}, pages 403--417, Cham. Springer
  International Publishing.

\bibitem[{Touvron et~al.(2023)Touvron, Lavril, Izacard, Martinet, Lachaux,
  Lacroix, Rozi{\`e}re, Goyal, Hambro, Azhar, Rodriguez, Joulin, Grave, and
  Lample}]{LLAMA}
Hugo Touvron, Thibaut Lavril, Gautier Izacard, Xavier Martinet, Marie-Anne
  Lachaux, Timoth{\'e}e Lacroix, Baptiste Rozi{\`e}re, Naman Goyal, Eric
  Hambro, Faisal Azhar, Aurelien Rodriguez, Armand Joulin, Edouard Grave, and
  Guillaume Lample. 2023.
\newblock {LLAMA}: Open and efficient foundation language models.
\newblock \emph{arXiv preprint arXiv:2302.13971}.

\bibitem[{Wang et~al.(2019)Wang, Pruksachatkun, Nangia, Singh, Michael, Hill,
  Levy, and Bowman}]{wang2019superglue}
Alex Wang, Yada Pruksachatkun, Nikita Nangia, Amanpreet Singh, Julian Michael,
  Felix Hill, Omer Levy, and Samuel Bowman. 2019.
\newblock Superglue: A stickier benchmark for general-purpose language
  understanding systems.
\newblock \emph{Advances in neural information processing systems}, 32.

\bibitem[{Wang et~al.(2020)Wang, K, Mayhew, and Roth}]{wang2020extending}
Zihan Wang, Karthikeyan K, Stephen Mayhew, and Dan Roth. 2020.
\newblock \href {https://doi.org/10.18653/v1/2020.findings-emnlp.240}
  {Extending multilingual {BERT} to low-resource languages}.
\newblock In \emph{Findings of the Association for Computational Linguistics:
  EMNLP 2020}, pages 2649--2656, Online. Association for Computational
  Linguistics.

\bibitem[{Wenzek et~al.(2020)Wenzek, Lachaux, Conneau, Chaudhary, Guzm{\'a}n,
  Joulin, and Grave}]{wenzek-etal-2020-ccnet}
Guillaume Wenzek, Marie-Anne Lachaux, Alexis Conneau, Vishrav Chaudhary,
  Francisco Guzm{\'a}n, Armand Joulin, and Edouard Grave. 2020.
\newblock \href {https://aclanthology.org/2020.lrec-1.494} {{CCN}et: Extracting
  high quality monolingual datasets from web crawl data}.
\newblock In \emph{Proceedings of the Twelfth Language Resources and Evaluation
  Conference}, pages 4003--4012, Marseille, France. European Language Resources
  Association.

\bibitem[{Wolf et~al.(2020)Wolf, Debut, Sanh, Chaumond, Delangue, Moi, Cistac,
  Rault, Louf, Funtowicz, Davison, Shleifer, von Platen, Ma, Jernite, Plu, Xu,
  Scao, Gugger, Drame, Lhoest, and Rush}]{wolf-etal-2020-transformers}
Thomas Wolf, Lysandre Debut, Victor Sanh, Julien Chaumond, Clement Delangue,
  Anthony Moi, Pierric Cistac, Tim Rault, Rémi Louf, Morgan Funtowicz, Joe
  Davison, Sam Shleifer, Patrick von Platen, Clara Ma, Yacine Jernite, Julien
  Plu, Canwen Xu, Teven~Le Scao, Sylvain Gugger, Mariama Drame, Quentin Lhoest,
  and Alexander~M. Rush. 2020.
\newblock \href {https://www.aclweb.org/anthology/2020.emnlp-demos.6}
  {Transformers: State-of-the-art natural language processing}.
\newblock In \emph{Proceedings of the 2020 Conference on Empirical Methods in
  Natural Language Processing: System Demonstrations}, pages 38--45, Online.
  Association for Computational Linguistics.

\bibitem[{Xue et~al.(2021)Xue, Constant, Roberts, Kale, Al-Rfou, Siddhant,
  Barua, and Raffel}]{mc4}
Linting Xue, Noah Constant, Adam Roberts, Mihir Kale, Rami Al-Rfou, Aditya
  Siddhant, Aditya Barua, and Colin Raffel. 2021.
\newblock \href {https://doi.org/10.18653/v1/2021.naacl-main.41} {m{T}5: A
  massively multilingual pre-trained text-to-text transformer}.
\newblock In \emph{Proceedings of the 2021 Conference of the North American
  Chapter of the Association for Computational Linguistics: Human Language
  Technologies}, pages 483--498, Online. Association for Computational
  Linguistics.

\bibitem[{Zhang et~al.(2022)Zhang, Roller, Goyal, Artetxe, Chen, Chen, Dewan,
  Diab, Li, Lin, Mihaylov, Ott, Shleifer, Shuster, Simig, Koura, Sridhar, Wang,
  and Zettlemoyer}]{OPT}
Susan Zhang, Stephen Roller, Naman Goyal, Mikel Artetxe, Moya Chen, Shuohui
  Chen, Christopher Dewan, Mona Diab, Xian Li, Xi~Victoria Lin, Todor Mihaylov,
  Myle Ott, Sam Shleifer, Kurt Shuster, Daniel Simig, Punit~Singh Koura, Anjali
  Sridhar, Tianlu Wang, and Luke Zettlemoyer. 2022.
\newblock {OPT}: Open pre-trained transformer language models.
\newblock \emph{arXiv preprint arXiv:2205.01068}.

\end{thebibliography}

\section{Language Resource References}
\label{lr:ref}
\bibliographystylelanguageresource{lrec-coling2024-natbib}
\bibliographylanguageresource{languageresource}

\appendix
\section{Appendix}

\subsection{Annotation details}

The specific annotation examples that were used to instruct all annotators are shown in Table~\ref{tab:examples_mono}. Since all annotators were also fluent English speakers, we opted to show the annotation examples as English texts. The main advantage of this approach is that the annotation instructions were the exact same across languages.

\subsection{Evaluation details}

The specific train, development and test set sizes per language and task are reported in Table~\ref{tab:fine_size}. Specific hyper-parameter settings (that differ from the default) are shown in Table~\ref{tab:fine_hyper}. For NER and POS, we use the NERmodel implementation of the Simpletransformers package \citep{simpletransformers}. For COPA, we use the ModelForMultipleChoice from the Transformers \citep{wolf-etal-2020-transformers} library. For CB, we use the SequenceClassification model from the same library.

\paragraph{COPA} For the COPA task, the training is not always stable. There are often runs where the training loss simply does not go down. These are considered to be failed runs. Failed runs are simply discarded, i.e., when averaging over 20 runs we do not take the failed runs into account.

\begin{table}[!htb]
    \centering
    \begin{tabular}{lrrr}
         \toprule
       &   \textbf{Train} & \textbf{Dev} & \textbf{Test} \\
   \midrule
   \textbf{POS} & & & \\
    \tabspace Croatian & 6,914 & 960 & 1,136 \\
    \tabspace Serbian & 3,328 & 536 & 520\\
     \tabspace Albanian & 5,307 & 611 & 708\\
      \tabspace Icelandic & 8,896 & 4,865 & 5,157\\
       \tabspace Slovenian & 10,903 & 1,250 & 1,282\\
      \midrule
      \textbf{NER} & & & \\
    \tabspace Croatian & 19,792 & 2,487 & 2,487\\
    \tabspace Serbian & 3,329 & 537 & 521\\
     \tabspace Albanian & 5,000 & 1,000 & 1,000\\
      \tabspace Icelandic & 10,651 & 6,479 & 5,889\\
       \tabspace Slovenian & 7,625 & 921 & 942\\
       \midrule
       \textbf{COPA} & 400 & 100 & 500 \\
       \midrule
       \textbf{CB} & 250 & 56 & 250 \\ 
         \bottomrule
    \end{tabular}
    \caption{\label{tab:fine_size}Train, development and test set sizes for the four tasks used during our automatic evaluation. For COPA and CB, the sizes are the same across languages.}
\end{table}

\begin{table}[!htb]
    \centering
    \begin{tabular}{lrrrr}
         \toprule
       & \textbf{POS} & \textbf{NER} & \textbf{COPA} & \textbf{CB} \\
   \midrule
   Learning rate & 1e-05 & 1e-05 & 1e-05 & 3e-05 \\
   Batch size & 8 & 8 & 8 & 4\\
   Epochs & 8 & 8 & 15 & 10 \\
   Max length & 512 & 512 & 100 & 512 \\
   Runs & 10 & 10 & 20 & 30 \\
         \bottomrule
    \end{tabular}
    
    \caption{\label{tab:fine_hyper}Hyper-parameter settings used across our evaluation experiments. Settings not mentioned are left at default.}
\end{table}

\begin{table*}[!htb]
    \centering
    \setlength{\tabcolsep}{4pt}
    \resizebox{\textwidth}{!}{
    \begin{tabular}{l}
         \toprule
       \textbf{Wrong language / Not language} \\ 
       \tspace \textbf{1:} STRAND 240 242 ECO:0000244|PDB:1FMK. \\
        \tspace \textbf{2:} box-shadow: 1px 1px 3px 2px \#121e03; \\
       \tspace \textbf{3:}  Ísland er fallegt land til að heimsækja.\\
   \midrule
   \textbf{Not running text} \\ 
       \tspace \textbf{1:} Archives Select Month December 2022 (2) November 2022 (11) October 2022 (14) September 2022  [...] \\
\tspace \textbf{2:} \#fish \#koi \#carpe \#carpekoi \#poisson \#japon \\
\tspace \textbf{3:} September 23, 2018 \\
\tspace \textbf{4:} Sheraton Signature Sleep Experience® | Sheraton Tirana Hotel | Official Website \\

   \midrule
   \textbf{Partially running text} \\    
    \tspace   \textbf{1:} bomb blew up in her face on Christmas Eve. Police refused to speculate about the \\
\tspace \textbf{2:} in approximately 13,000 women every year in the United States, and kills almost 5,000 American \\
\tspace \textbf{3:} Complete Your Bachelors Degree or Associate Degree | Charter …  \\
\tspace \textbf{4:} The premium is the amount you'll pay for the huge benefits protected  \\

   \midrule
   \textbf{Running text, but (slightly) non-standard} \\ 
      \tspace \textbf{1:} What The Riga Elections Say About Latvian Politics – Analysis – Eurasia Review  \\
   \tspace \textbf{2:} Demonstrated critical thinking and decision-making competencies  \\
    \tspace \textbf{3:} Friday.4th: At Noon, the Detachment of Marines fired 3 vollies in honour of the Day.  \\
  \tspace  \textbf{4:} HOW DO I WRITE A BUSINESS REPORT?  \\
  \tspace  \textbf{5:} (c) It is because of God, then, that we have language and words \\
  \tspace  \textbf{6:} I get a long line of numbers (where you can extract the windspeed from), but the outcome is strange. \\
    \hspace{0.55cm}  This is shown in my log file:  \\
   \midrule
   \textbf{Publishable text} \\ 
   \tspace \textbf{1:} You don't mean that, said Bones hoarsely.  \\
   \tspace \textbf{2:} Sounds delicious. My daughter makes it with a puréed jalapeño swirl that looks and tastes amazing.  \\
  \tspace  \textbf{3:} Does your business need an interactive website or app?  \\
  \tspace  \textbf{4:} The Caucasian rugs are made in the regions located in the mountain chain of the Caucasus, an area situated \\
  \hspace{0.55cm} between the Black Sea and the Caspian Sea. The area is spanned across Georgia, Russian, Armenia and Azerbaijan. \\

         \bottomrule
    \end{tabular}
    }
    \caption{\label{tab:examples_mono}Example English texts for each annotation category that were given to all annotators. All examples are actual examples taken from the English versions of the respective corpora.}
\end{table*}

\end{document}